\documentclass{article}
\usepackage{times}
\usepackage{graphicx} 
\usepackage{subfigure} 
\usepackage{natbib}
\usepackage{algorithm}
\usepackage{algorithmic}
\usepackage{hyperref}
\usepackage{amssymb}
\usepackage{amsmath}
\usepackage{mathtools, xparse}

\usepackage[accepted]{icml2017}
\usepackage{paralist}
\usepackage{comment}
\specialcomment{rahul}
 {\begingroup\color{blue}}{\endgroup}

\icmltitlerunning{Robust Adversarial Reinforcement Learning}

\begin{document} 
\twocolumn[
\icmltitle{Robust Adversarial Reinforcement Learning}




\begin{icmlauthorlist}
\icmlauthor{Lerrel Pinto}{cmu}
\icmlauthor{James Davidson}{goo_br}
\icmlauthor{Rahul Sukthankar}{goo_re}
\icmlauthor{Abhinav Gupta}{cmu,goo_re}
\end{icmlauthorlist}

\icmlaffiliation{cmu}{Carnegie Mellon University}
\icmlaffiliation{goo_br}{Google Brain}
\icmlaffiliation{goo_re}{Google Research}

\icmlcorrespondingauthor{Lerrel Pinto}{lerrelp@cs.cmu.edu}

\icmlkeywords{Robust Learning, Reinforcement Learning, Adversarial Agents}

\vskip 0.3in
]

\printAffiliationsAndNotice{}

\begin{abstract} 
Deep neural networks coupled with fast simulation and improved computation have led to recent successes in the field of reinforcement learning (RL). However, most current RL-based approaches fail to generalize since: (a) the gap between simulation and real world is so large that policy-learning approaches fail to transfer; (b) even if policy learning is done in real world, the data scarcity leads to failed generalization from training to test scenarios (e.g., due to different friction or object masses). Inspired from $H_{\infty}$ control methods, we note that both modeling errors and differences in training and test scenarios can be viewed as extra forces/disturbances in the system. This paper proposes the idea of robust adversarial reinforcement learning (RARL), where we train an agent to operate in the presence of a destabilizing adversary that applies disturbance forces to the system. The jointly trained adversary is reinforced -- that is, it learns an optimal destabilization policy. 
We formulate the policy learning as a zero-sum, minimax objective function. Extensive experiments in multiple environments (InvertedPendulum, HalfCheetah, Swimmer, Hopper and Walker2d) conclusively demonstrate that our method  (a) improves training stability;  (b) is robust to differences in training/test conditions; and c) outperform the baseline even in the absence of the adversary.
\end{abstract} 
\section{Introduction}

\begin{figure*}[t!]
\begin{center}
\includegraphics[width=6.5in]{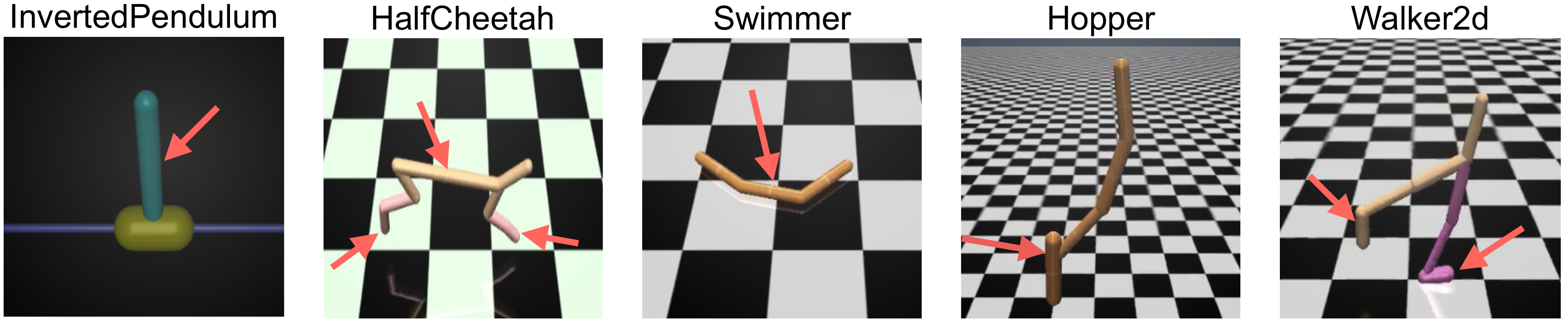}
\end{center}
\caption{We evaluate RARL on a variety of OpenAI gym problems. The adversary learns to apply destabilizing forces on specific points (denoted by red arrows) on the system, encouraging the protagonist to learn a robust control policy. These policies also transfer better to new test environments, with different environmental conditions and where the adversary may or may not be present.
}
\label{fig:all_envs}
\end{figure*}

High-capacity function approximators such as deep neural networks have led to increased success in the field of reinforcement learning~\cite{mnih2015human,silver2016mastering, gu2016continuous, lillicrap2015continuous, mordatch2015interactive}. However, a major bottleneck for such policy-learning methods is their reliance on data: training high-capacity models requires huge amounts of training data/trajectories. While this training data can be easily obtained for tasks like games (e.g., Doom, Montezuma's Revenge)~\cite{mnih2015human}, data-collection and policy learning for real-world physical tasks are significantly more challenging. 

There are two possible ways to perform policy learning for real-world physical tasks:
\begin{compactitem}
\item {\bf Real-world Policy Learning:} The first approach is to learn the agent's policy in the real-world. However, training in the real-world is too expensive, dangerous and time-intensive leading to scarcity of data. Due to scarcity of data, training is often restricted to a limited set of training scenarios, causing overfitting. If the test scenario is different (e.g., different friction coefficient), the learned policy fails to generalize. Therefore, we need a learned policy that is robust and generalizes well across a range of scenarios.

\item {\bf Learning in simulation:} One way of escaping the data scarcity in the real-world is to transfer a policy learned in a simulator to the real world. However the environment and physics of the simulator are not exactly the same as the real world. This reality gap often results in unsuccessful transfer if the learned policy isn't robust to modeling errors~\cite{christiano2016transfer,rusu2016sim}. 
\end{compactitem}

Both the test-generalization and simulation-transfer issues are further exacerbated by the fact that many policy-learning algorithms are stochastic in nature. For many hard physical tasks such as Walker2D~\cite{openaigym}, only
a small fraction of runs
leads to stable walking policies. This makes these approaches even more time and data-intensive. What we need is an approach that is significantly more stable/robust in learning policies across different runs and initializations while requiring less data during training.

So, how can we model uncertainties and learn a policy robust to all uncertainties? How can we model the gap between simulations and real-world? We begin with the insight that modeling errors can be viewed as extra forces/disturbances in the system~\cite{basar2008h}. For example, high friction at test time might be modeled as extra forces at contact points against the direction of motion. Inspired by this observation, this paper proposes the idea of modeling uncertainties via an adversarial agent that applies disturbance forces to the system. Moreover, the adversary is reinforced -- that is, it learns an optimal policy to thwart the original agent's goal. Our proposed method, Robust Adversarial Reinforcement Learning (\textsc{RARL}), jointly trains a pair of agents, a protagonist and an adversary, where the protagonist learns to fulfil the original task goals while being robust to the disruptions generated by its adversary.


We perform extensive experiments to evaluate \textsc{RARL} on multiple OpenAI gym environments like InvertedPendulum, HalfCheetah, Swimmer, Hopper and Walker2d (see Figure~\ref{fig:all_envs}). We demonstrate that our proposed approach is: {\bf (a) Robust to model initializations:} The learned policy performs better given different model parameter initializations and random seeds. This alleviates the data scarcity issue by reducing sensitivity of learning. {\bf (b) Robust to modeling errors and uncertainties:} The learned policy generalizes significantly better to different test environment settings (e.g., with different mass and friction values). 

\subsection{Overview of RARL}
Our goal is to learn a policy that is robust to modeling errors in simulation or mismatch between training and test scenarios. For example, we would like to learn policy for Walker2D that works not only on carpet (training scenario) but also generalizes to walking on ice (test scenario). Similarly, other parameters such as the mass of the walker might vary during training and test. One possibility is to list all such parameters (mass, friction etc.) and learn an ensemble of policies for different possible variations~\cite{Rajeswaran2016epopt}. 
But explicit consideration of all possible 
parameters of how simulation and real world might differ or what parameters can change between training/test
is infeasible.

Our core idea is to model the differences during training and test scenarios via extra forces/disturbances in the system. Our hypothesis is that if we can learn a policy that is robust to all disturbances, then this policy will be robust to changes in training/test situations; and hence generalize well. But is it possible to sample trajectories under all possible disturbances? In unconstrained scenarios, the space of possible disturbances could be larger than the space of possible actions, which makes sampled trajectories even sparser in the joint space. 

To overcome this problem, we advocate a two-pronged approach:

{\bf (a) Adversarial agents for modeling disturbances:} Instead of sampling all possible disturbances, we jointly train a second agent
(termed the adversary), whose goal is to impede the original agent (termed the protagonist) by applying destabilizing forces.
The adversary is rewarded only for the failure of the protagonist. Therefore, the adversary learns to sample hard examples: disturbances which will make original agent fail; the protagonist learns a policy that is robust to any disturbances created by the adversary.

\noindent {\bf (b) Adversaries that incorporate domain knowledge:} The naive way of developing an adversary would be to simply give it the same action space as the protagonist -- like a driving student and driving instructor fighting for control of a dual-control car.  However, our proposed approach is much richer and is not limited to symmetric action spaces -- we can exploit domain knowledge to: focus the adversary on the protagonist's weak points; and since the adversary is in a simulated environment, we can give the adversary ``super-powers'' -- the ability to affect the robot or environment in ways the protagonist cannot (e.g., suddenly change a physical parameter like frictional coefficient or mass).



\section{Background}
Before we delve into the details of RARL, we first outline our terminology, standard reinforcement learning setting and two-player zero-sum games from which our paper is inspired.

\subsection{Standard reinforcement learning on MDPs}
In this paper we examine continuous space MDPs that are represented by the tuple: $(\mathcal{S},\mathcal{A},\mathcal{P},r,\gamma, s_0)$, where $\mathcal{S}$ is a set of continuous states and $\mathcal{A}$ is a set of continuous actions, $\mathcal{P}: \mathcal{S} \times \mathcal{A} \times \mathcal{S} \rightarrow \mathbb{R}$ is the transition probability, $r: \mathcal{S} \times \mathcal{A} \rightarrow \mathbb{R}$ is the reward function, $\gamma$ is the discount factor, and $s_0$ is the initial state distribution. 

Batch policy algorithms like~\cite{williams1992simple,kakade2002natural,schulman2015trust} attempt to learn a stochastic policy $\pi_\theta:\mathcal{S} \times \mathcal{A} \rightarrow \mathbb{R}$ that maximizes the cumulative discounted reward $\sum_{t=0}^{T-1} \gamma^{t}r(s_t,a_t)$. Here, $\theta$ denotes the parameters for the policy $\pi$ which takes action $a_t$ given state $s_t$ at timestep $t$.

\subsection{Two-player zero-sum discounted games}
The adversarial setting we propose can be expressed as a two player $\gamma$ discounted zero-sum Markov game~\cite{littman1994markov, scherrer2015approximate}. This game MDP can be expressed as the tuple: $(\mathcal{S},\mathcal{A}_1,\mathcal{A}_2, \mathcal{P},r,\gamma, s_0)$ where $\mathcal{A}_1$ and $\mathcal{A}_2$ are the continuous set of actions the players can take. $\mathcal{P}: \mathcal{S} \times \mathcal{A}_1 \times \mathcal{A}_2 \times \mathcal{S} \rightarrow \mathbb{R}$ is the transition probability density and $r: \mathcal{S} \times \mathcal{A}_1 \times \mathcal{A}_2 \rightarrow \mathbb{R}$ is the reward of both players. If player 1 (protagonist) is playing strategy $\mu$ and player 2 (adversary) is playing the strategy $\nu$, the reward function is $r_{\mu,\nu}=E_{a^1\sim \mu(.|s),a^2\sim \nu(.|s)}[r(s,a^1,a^2)]$. A zero-sum two-player game can be seen as player 1 maximizing the $\gamma$ discounted reward while player 2 is minimizing it.



\section{Robust Adversarial RL}

\subsection{Robust Control via Adversarial Agents}
Our goal is to learn the policy of the protagonist (denoted by $\mu$) such that it is better (higher reward) and robust (generalizes better to variations in test settings). In the standard reinforcement learning setting, for a given transition function $\mathcal{P}$, we can learn policy parameters $\theta^\mu$ such that the expected reward is maximized where expected reward for policy $\mu$ from the start $s_0$ is
\begin{equation}
    \rho(\mu; \theta^\mu,\mathcal{P}) = \mathop{\mathbb{E}}\left[\sum_{t=0}^{T} \gamma^t r(s_t,a_t) | s_0, \mu, \mathcal{P}\right] .
\end{equation}
Note that in this formulation the expected reward is conditioned on the transition function since the the transition function defines the roll-out of states. In standard-RL settings, the transition function is fixed (since the physics engine and parameters such as mass, friction are fixed). However, in our setting, we assume that the transition function will have modeling errors and that there will be differences between training and test conditions. Therefore, in our general setting, we should estimate policy parameters $\theta^\mu$ such that we maximize the expected reward over different possible transition functions as well. Therefore,
\begin{equation}
   \rho(\mu; \theta^\mu)  = \mathop{\mathbb{E}}_\mathcal{P}\left[\mathop{\mathbb{E}} \left[\sum_{t=0}^{T} \gamma^t r(s_t,a_t) | s_0, \mu,  \mathcal{P}\right]\right] .
\end{equation}
Optimizing for the expected reward over \emph{all} transition functions optimizes \emph{mean performance}, which is a risk neutral formulation that assumes a known distribution over model parameters. A large fraction of policies learned under such a formulation are likely to fail in a different environment.
Instead, inspired by work in robust control~\cite{tamar2014optimizing,Rajeswaran2016epopt}, we choose to optimize for conditional value at risk (CVaR):
\begin{equation}
     \rho_{RC} = \mathop{\mathbb{E}}\left[ \rho | \rho \leq \mathcal{Q}_\alpha(\rho)\right]
\end{equation}
where $\mathcal{Q}_\alpha(\rho)$ is the $\alpha$-quantile of $\rho$-values. Intuitively, in robust control, we want to maximize the worst-possible $\rho$-values. But how do you tractably sample trajectories that are in worst $\alpha$-percentile? Approaches like EP-Opt~\cite{Rajeswaran2016epopt} sample these worst percentile trajectories by changing parameters such as friction, mass of objects, etc.\ during rollouts. 

Instead, we introduce an adversarial agent that applies forces on pre-defined locations, and this agent tries to change the trajectories such that reward of the protagonist is minimized. Note that since the adversary tries to minimize the protagonist's reward, it ends up sampling trajectories from worst-percentile leading to robust control-learning for the protagonist. If the adversary is kept fixed, the protagonist could learn to overfit to its adversarial actions. Therefore, instead of using either a random or a fixed-adversary, we advocate generating the adversarial actions using a learned policy $\nu$. We would also like to point out the connection between our proposed approach and the practice of hard-example mining~\cite{Sung94, ShrivastavaGG16}. The adversary in RARL learns to sample hard-examples (worst-trajectories) for the protagonist to learn. Finally, instead of using $\alpha$ as percentile-parameter, RARL is parameterized by the magnitude of force available to the adversary. As the adversary becomes stronger, RARL optimizes for lower percentiles. However, very high magnitude forces lead to very biased sampling and make the learning unstable. In the extreme case, an unreasonably strong adversary can always prevent the protagonist from achieving the task. Analogously, the traditional RL baseline is equivalent to training with an impotent (zero strength) adversary.



\subsection{Formulating Adversarial Reinforcement Learning}
In our adversarial game, at every timestep $t$ both players observe the state $s_t$ and take actions $a^1_t \sim \mu(s_t)$ and $a^2_t \sim \nu(s_t)$. The state transitions $s_{t+1} = \mathcal{P}(s_t, a^1_t, a^2_t)$ and a reward $r_t = r(s_t, a^1_t, a^2_t)$ is obtained from the environment. In our zero-sum game, the protagonist gets a reward $r^1_t = r_t$ while the adversary gets a reward $r^2_t = -r_t$. Hence each step of this MDP can be represented as $(s_t, a^1_t, a^2_t, r^1_t, r^2_t, s_{t+1})$.

The protagonist seeks to maximize the following reward function,
\begin{equation}
\label{eq:reward}
    R^1=E_{s_0\sim\rho,a^1\sim \mu(s),a^2\sim \nu(s)}[\sum_{t=0}^{T-1} r^1(s,a^1,a^2)] .
\end{equation}
Since, the policies $\mu$ and $\nu$ are the only learnable components, $R^1 \equiv R^1(\mu,\nu)$.
%
Similarly the adversary attempts to maximize its own reward: $R^2 \equiv R^2(\mu,\nu) = -R^1(\mu,\nu)$.
%
One way to solve this MDP game is by discretizing the continuous state and action spaces and using dynamic programming to solve. \cite{scherrer2015approximate,patek1997stochastic} show that notions of minimax equilibrium and Nash equilibrium are equivalent for this game with optimal equilibrium reward:
\begin{equation}
    R^{1*} = \min_{\nu}\max_{\mu} R^1(\mu,\nu) = \max_{\mu}\min_{\nu} R^1(\mu,\nu)
\end{equation}
However solutions to finding the Nash equilibria strategies often involve greedily solving $N$ minimax equilibria for a zero-sum matrix game, with $N$ equal to the number of observed datapoints. The complexity of this greedy solution is exponential in the cardinality of the action spaces, which makes it prohibitive~\cite{scherrer2015approximate}.

Most Markov Game approaches require solving for the equilibrium solution for a multiplayer value or minimax-Q function at each iteration. This requires evaluating a typically intractable minimax optimization problem.  Instead, we focus on learning stationary policies $\mu^*$ and $\nu^*$ such that $R^1(\mu^*,\nu^*) \rightarrow R^{1*}$. This way we can avoid this costly optimization at each iteration as we just need to approximate the advantage function and not determine the equilibrium solution at each iteration.

\subsection{Proposed Method: RARL}
Our algorithm (\textsc{RARL}) optimizes both of the agents using the following alternating procedure. In the first phase, we learn the protagonist's policy while holding the adversary's policy fixed. Next, the protagonist's policy is held constant and the adversary's policy is learned. This sequence is repeated until convergence.


Algorithm~\ref{alg:adv_rl} outlines our approach in detail. The initial parameters for both players' policies are sampled from a random distribution. In each of the $N_\text{iter}$ iterations, we carry out a two-step (alternating) optimization procedure. First, for $N_{\mu}$ iterations, the parameters of the adversary $\theta^{\nu}$ are held constant while the parameters $\theta^{\mu}$ of the protagonist are optimized to maximize $R^1$ (Equation~\ref{eq:reward}). The \emph{roll} function samples $N_\text{traj}$ trajectories given the environment definition $\mathcal{E}$ and the policies for both the players. Note that $\mathcal{E}$ contains the transition function $\mathcal{P}$ and the reward functions $r^1$ and $r^2$ to generate the trajectories. The $t^\text{th}$ element of the $i^\text{th}$ trajectory is of the form $(s^i_t,a^{1i}_t,a^{2i}_t,r^{1i}_t,r^{2i}_t)$. 
These trajectories are then split such that the $t^\text{th}$ element of the $i^\text{th}$ trajectory is of the form $(s^i_t,a^{i}_t=a^{1i}_t,r^{i}_t=r^{1i}_t)$. The protagonist's parameters $\theta^{\mu}$ are then optimized using a policy optimizer. For the second step, player 1's parameters $\theta^{\mu}$ are held constant for the next $N_{\nu}$ iterations. $N_\text{traj}$ Trajectories are sampled and split into trajectories such that $t^\text{th}$ element of the $i^\text{th}$ trajectory is of the form $(s^i_t,a^{i}_t=a^{2i}_t,r^{i}_t=r^{2i}_t)$. Player 2's parameters $\theta^{\nu}$ are then optimized. This alternating procedure is repeated for $N_\text{iter}$ iterations.

\begin{algorithm}[tb]
   \caption{RARL (proposed algorithm)}
   \label{alg:adv_rl}
\begin{algorithmic}
   \STATE {\bfseries Input:} Environment $\mathcal{E}$; Stochastic policies $\mu$ and $\nu$ 
   \STATE {\bfseries Initialize:} Learnable parameters $\theta^{\mu}_{0}$ for $\mu$ and $\theta^{\nu}_{0}$ for $\nu$
   \FOR {$i$=1,2,..$N_\text{iter}$}
   \STATE $\theta^{\mu}_{i} \leftarrow \theta^{\mu}_{i-1}$
   \FOR {$j$=1,2,..$N_{\mu}$}
   \STATE $\{(s^i_t,a^{1i}_t,a^{2i}_t,r^{1i}_t,r^{2i}_t)\} \leftarrow \text{roll}(\mathcal{E}, \mu_{\theta^{\mu}_{i}}, \nu_{\theta^{\nu}_{i-1}}, N_\text{traj})$
   \STATE $\theta^{\mu}_{i} \leftarrow \text{policyOptimizer}(\{(s^i_t,a^{1i}_t,r^{1i}_t)\}, \mu, \theta^{\mu}_{i})$
   \ENDFOR
   \STATE $\theta^{\nu}_{i} \leftarrow \theta^{\nu}_{i-1}$
   \FOR {$j$=1,2,..$N_{\nu}$}
   \STATE $\{(s^i_t,a^{1i}_t,a^{2i}_t,r^{1i}_t,r^{2i}_t)\} \leftarrow \text{roll}(\mathcal{E}, \mu_{\theta^{\mu}_{i}}, \nu_{\theta^{\nu}_{i}},N_\text{traj})$
   \STATE $\theta^{\nu}_{i} \leftarrow \text{policyOptimizer}(\{(s^i_t,a^{2i}_t,r^{2i}_t)\}, \nu, \theta^{\nu}_{i})$
   \ENDFOR
   
   \ENDFOR
   \STATE {\bf Return: }{$\theta^{\mu}_{N_\text{iter}}, \theta^{\nu}_{N_\text{iter}}$}
\end{algorithmic}
\end{algorithm}

\section{Experimental Evaluation}
We now demonstrate the robustness of the RARL algorithm: (a) for training with different initializations; (b) for testing with different conditions; (c) for adversarial disturbances 
in the testing environment. But first we will describe our implementation and test setting followed by evaluations and results of our algorithm.

\subsection{Implementation}
Our implementation of the adversarial environments build on OpenAI gym's~\cite{openaigym} control environments with the MuJoCo ~\cite{todorov2012mujoco} physics simulator. Details of the environments and their corresponding adversarial disturbances are (also see Figure~\ref{fig:all_envs}): 

\noindent {\bf InvertedPendulum:} The inverted pendulum is mounted on a pivot point on a cart, with the cart restricted to linear movement in a plane. The state space is 4D: position and velocity for both the cart and the pendulum. The protagonist can apply 1D forces to keep the pendulum upright. The adversary applies a 2D force on the center of pendulum in order to destabilize it.

\noindent {\bf HalfCheetah:} The half-cheetah is a planar biped robot with 8 rigid links, including two legs and a torso, along with 6 actuated joints. The 17D state space includes joint angles and joint velocities. The adversary applies a 6D action with 2D forces on the torso and both feet in order to destabilize it. 

\noindent {\bf Swimmer:} The swimmer is a planar robot with 3 links and 2 actuated joints in a viscous container, with the goal of moving forward. The 8D state space includes joint angles and joint velocities. The adversary applies a 3D force to the center of the swimmer.

\noindent {\bf Hopper:} The hopper is a planar monopod robot with 4 rigid links, corresponding to the torso, upper leg, lower leg, and foot, along with 3 actuated joints. The 11D state space includes joint angles and joint velocities. The adversary applies a 2D force on the foot.

\noindent {\bf Walker2D:} The walker is a planar biped robot consisting of 7 links, corresponding to two legs and a torso, along with 6 actuated joints. The 17D state space includes joint angles and joint velocities. The adversary applies a 4D action with 2D forces on both the feet.

Our implementation of \textsc{RARL} is built on top of rllab~\cite{duan2016benchmarking} and uses Trust Region Policy Optimization (TRPO)~\cite{schulman2015trust} as the policy optimizer. For all the tasks and for both the protagonist and adversary, we use a policy network with two hidden layers with 64 neurons each. We train both ~\textsc{RARL} and the baseline for 100 iterations on InvertedPendulum and for 500 iterations on the other tasks. Hyperparameters of TRPO are selected by grid search.

\subsection{Evaluating Learned Policies}
\begin{figure}[!h]
\begin{center}
\includegraphics[width=3.2in]{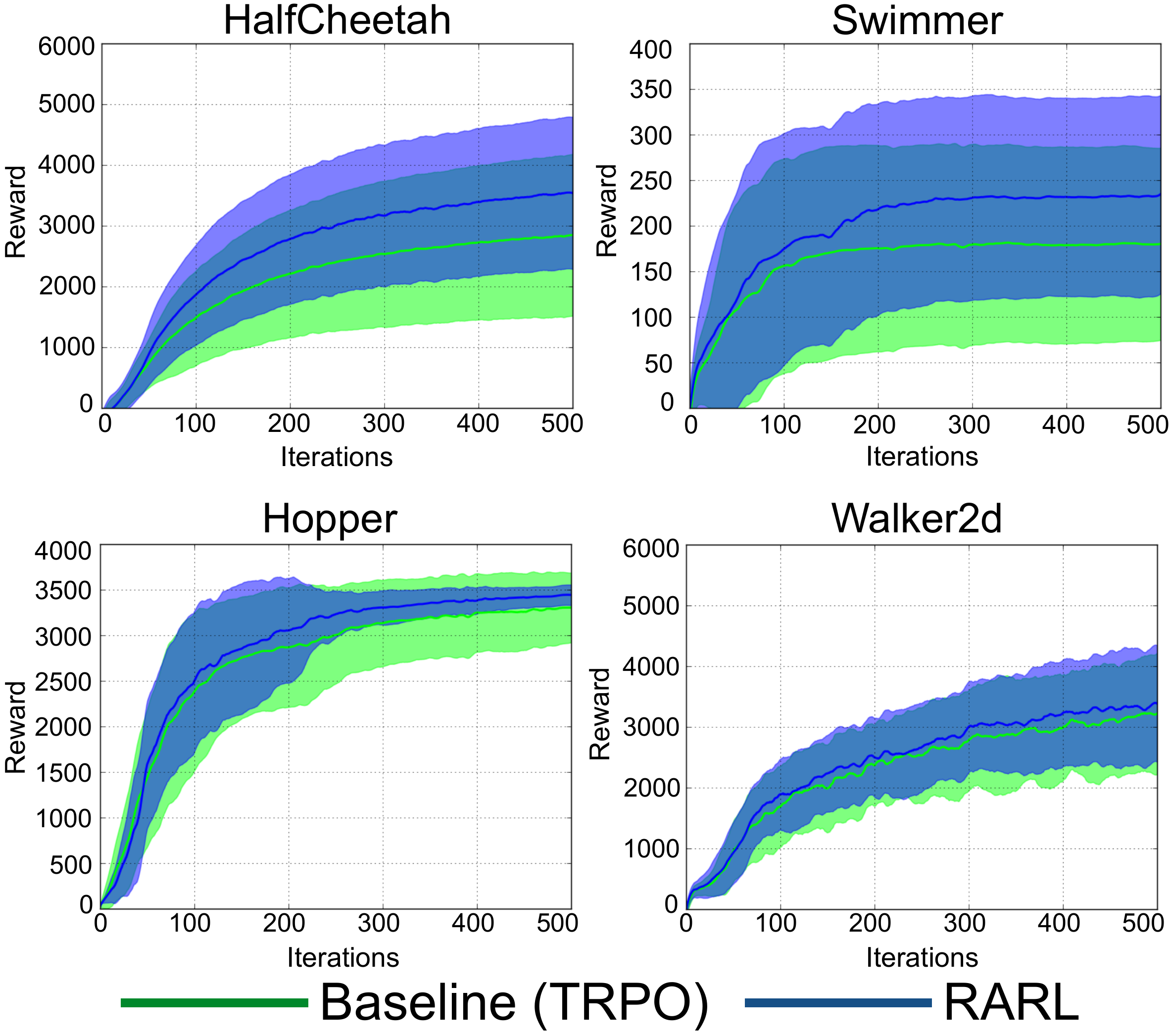}
\end{center}
\caption{Cumulative reward curves for \textsc{RARL} trained policies versus the baseline (TRPO) when tested without any disturbance. For all the tasks, \textsc{RARL} achieves a better mean than the baseline. For tasks like Hopper, we also see a significant reduction of variance across runs.}
\label{fig:all_comp}
\end{figure}

We evaluate the robustness of our RARL approach compared to the strong TRPO baseline. Since our policies are stochastic in nature and the starting state is also drawn from a distribution, we learn 50 policies for each task with different seeds/initializations. First, we report the mean and variance of cumulative reward (over 50 policies) as a function of the training iterations. Figure~\ref{fig:all_comp} shows the mean and variance of the rewards of learned policies for the task of HalfCheetah, Swimmer, Hopper and Walker2D. We omit the graph for InvertedPendulum because the task is easy and both TRPO and RARL show similar performance and similar rewards. As we can see from the figure, for all the four tasks RARL learns a better policy in terms of mean reward and variance as well. This clearly shows that the policy learned by RARL is better than the policy learned by TRPO even when there is no disturbance or change of settings between training and test conditions. Table~\ref{tab:best_comp} reports the average rewards with their standard deviations for the best learned policy.

\begin{table*}[t]
\centering
\caption{Comparison of the best policy learned by \textsc{RARL} and the baseline (mean$\pm$one standard deviation)}
\label{tab:best_comp}
\begin{tabular}{c|ccccc}
         & InvertedPendulum & HalfCheetah  & Swimmer    & Hopper      & Walker2d      \\ \hline
Baseline & $1000\pm0.0$     & $5093\pm44$ & $358\pm2.4$  & $3614\pm2.16$ & $5418\pm87$  \\
RARL    & $1000\pm0.0$     & $5444\pm97$ & $354\pm1.5$ & $3590\pm7.4$ & $5854\pm159$
\end{tabular}
\end{table*}

\begin{figure}[!h]
\begin{center}
\includegraphics[width=3.2in]{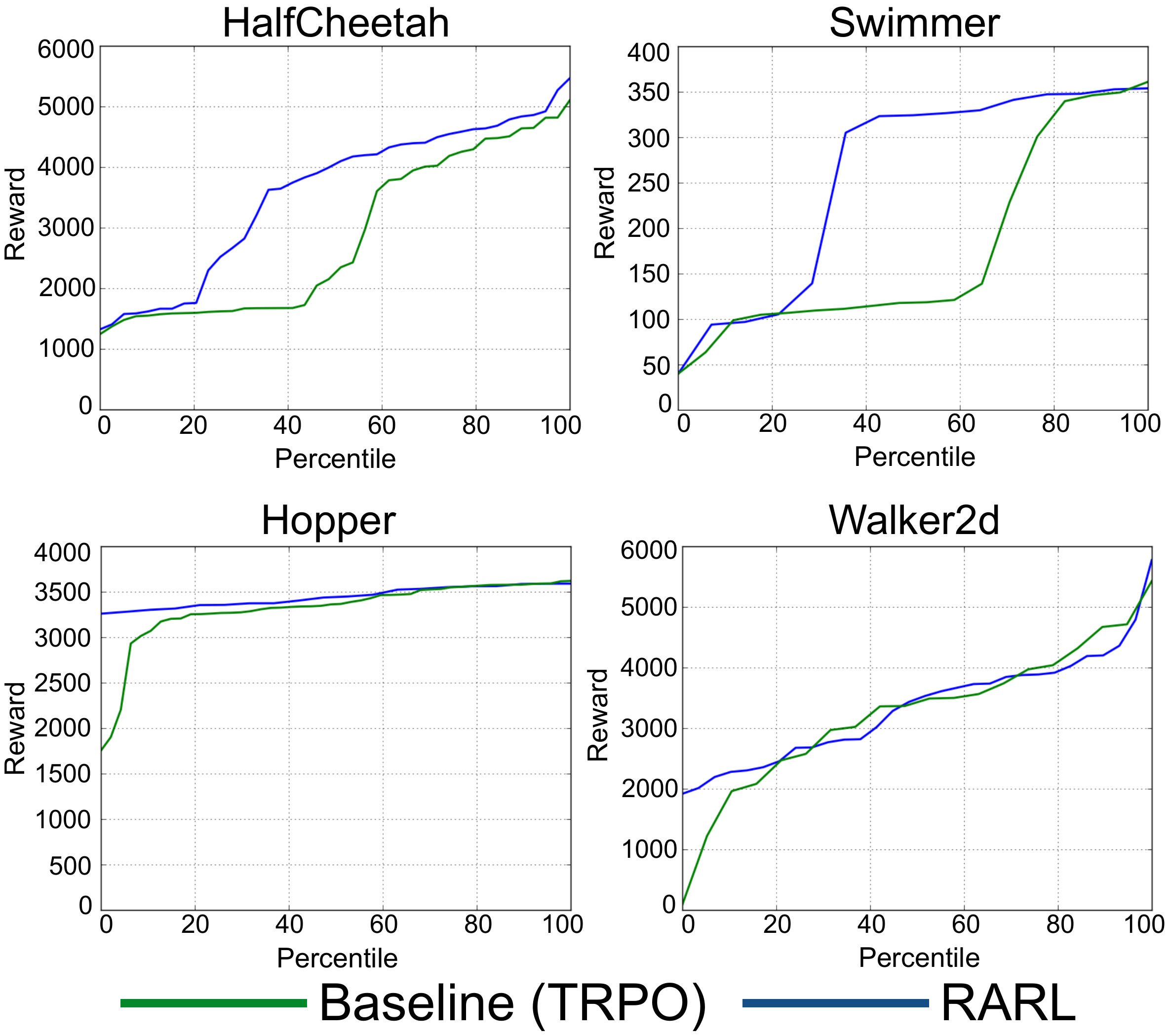}
\end{center}
\caption{We show percentile plots \textbf{without any disturbance} to show the robustness of \textsc{RARL} compared to the baseline. Here the algorithms are run on multiple initializations and then sorted to show the n$^{th}$ percentile of cumulative final reward.}
\label{fig:per_zero}
\end{figure}

However, the primary focus of this paper is to show robustness in training these control policies. One way of visualizing this is by plotting the average rewards for the n$^{th}$ percentile of trained policies. Figure~\ref{fig:per_zero} plots these percentile curves and highlight the significant gains in robustness for training for the HalfCheetah, Swimmer and Hopper tasks. 

\subsection{Robustness under Adversarial Disturbances}
\begin{figure}[!h]
\begin{center}
\includegraphics[width=3.2in]{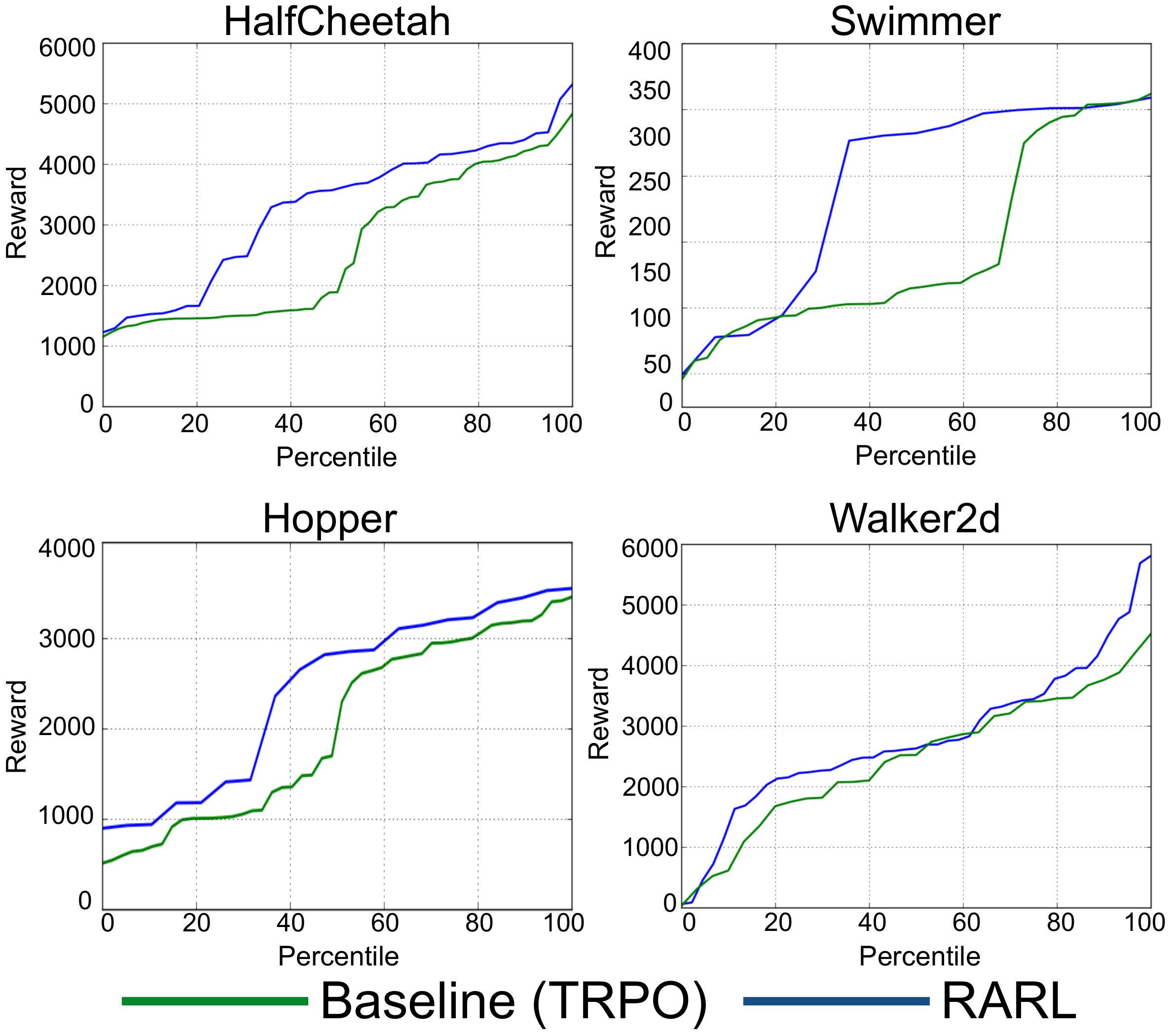}
\end{center}
\caption{Percentile plots with a \textbf{learned adversarial disturbance} show the robustness of \textsc{RARL} compared to the baseline in the presence of an adversary. Here the algorithms are run on multiple initializations followed by learning an adversarial disturbance that is applied at test time.}
\label{fig:per_adv}
\end{figure}

While deploying controllers in the real world, unmodeled environmental effects can cause controllers to fail. One way of measuring robustness to such effects is by measuring the performance of our learned control polices in the presence of an adversarial disturbance. For this purpose, we train an adversary to apply a disturbance while holding the protagonist's policy constant. We again show the percentile graphs as described in the section above. RARL's control policy, since it was trained on similar adversaries, performs better, as seen in Figure~\ref{fig:per_adv}.

\subsection{Robustness to Test Conditions}
Finally, we evaluate the robustness and generalization of the learned policy with respect to varying test conditions. In this section, we train the policy based on certain mass and friction values; however at test time we evaluate the policy when different mass and friction values are used in the environment. Note we omit evaluation of Swimmer since the policy for the swimming task is not significantly impacted by a change mass or friction.

\subsubsection{Evaluation with changing mass}
\begin{figure}[!h]
\begin{center}
\includegraphics[width=3.2in]{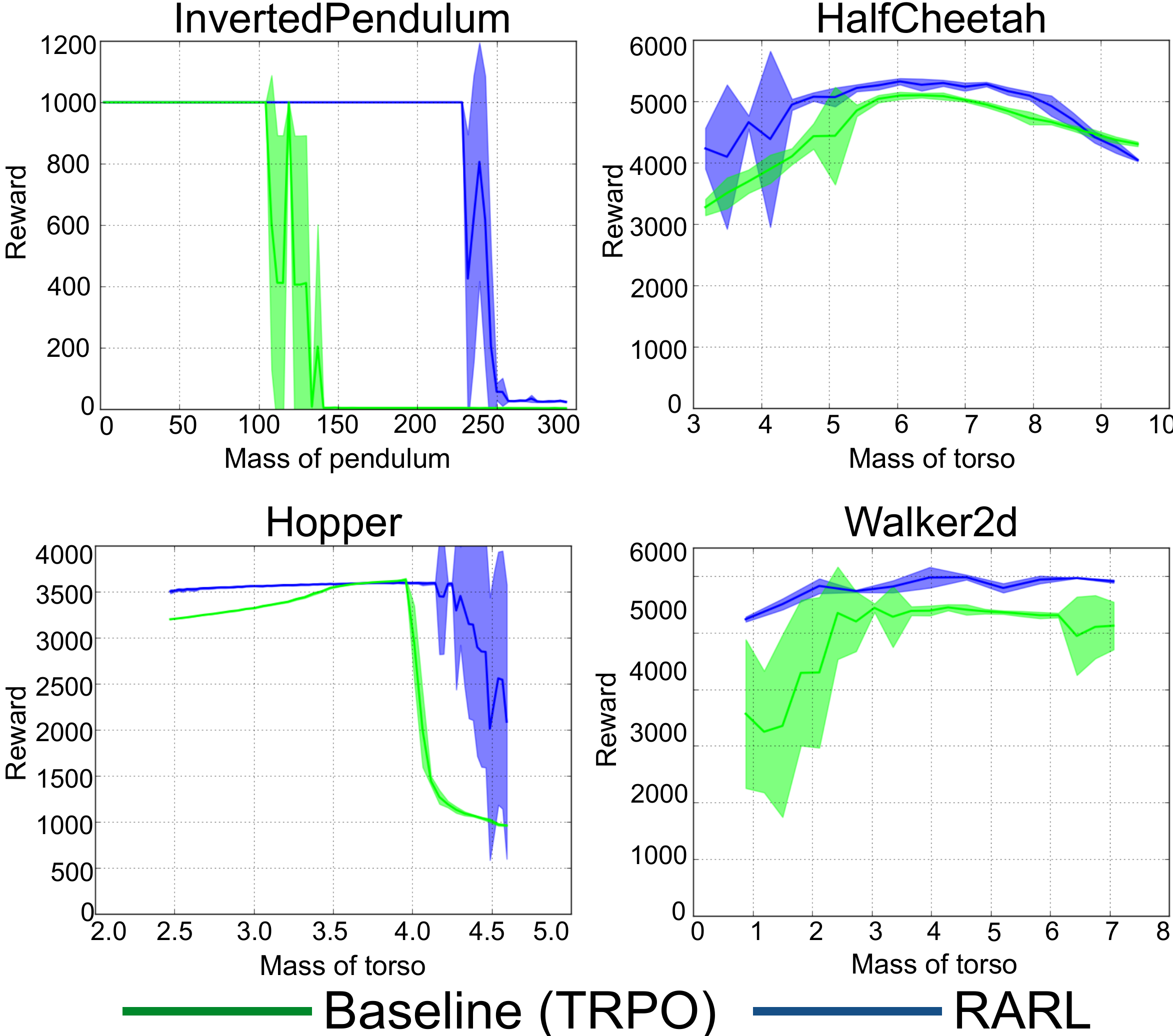}
\end{center}
\caption{The graphs show robustness of \textsc{RARL} policies to changing mass between training and testing. For the InvertedPendulum the mass of the pendulum is varied, while for the other tasks, the mass of the torso is varied.}
\label{fig:rob_mass}
\end{figure}

We describe the results of training with the standard mass variables in OpenAI gym while testing it with different mass. Specifically, the mass of InvertedPendulum, HalfCheetah, Hopper and Walker2D were 4.89, 6.36, 3.53 and 3.53 respectively. At test time, we evaluated the learned policies by changing mass values and estimating the average cumulative rewards. Figure~\ref{fig:rob_mass} plots the average rewards and their standard deviations against a given torso mass (horizontal axis). As seen in these graphs, RARL policies generalize significantly better. 

\subsubsection{Evaluation with changing friction}

\begin{figure*}[t!]
\begin{center}
\includegraphics[width=5.5in]{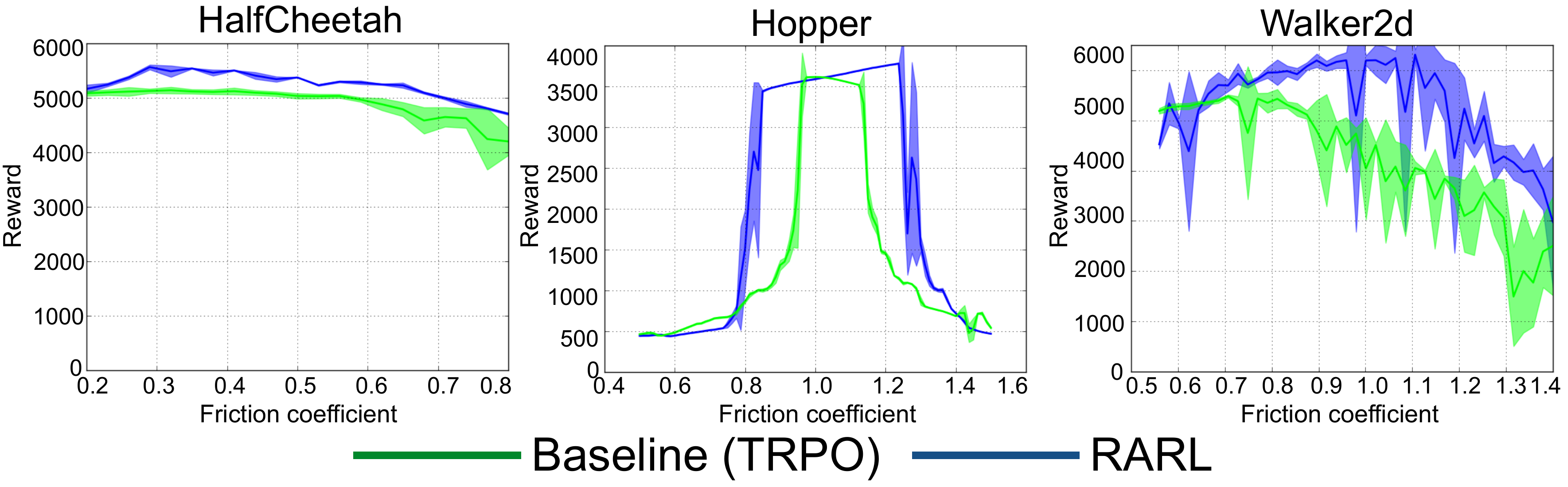}
\end{center}
\caption{The graphs show robustness of \textsc{RARL}  policies to changing friction between training and testing. Note that we exclude the results of InvertedPendulum and the Swimmer because friction is not relevant to those tasks.}
\label{fig:rob_fric}
\end{figure*}

Since several of the control tasks involve contacts and friction (which is often poorly modeled), we evaluate robustness to different friction coefficients in testing. Similar to the evaluation of robustness to mass, the model is trained with the standard variables in OpenAI gym. Figure~\ref{fig:rob_fric} shows the average reward values with different friction coefficients at test time. It can be seen that the baseline policies fail to generalize and the performance falls significantly when the test friction is different from training. On the other hand RARL shows more resilience to changing friction values.

\begin{figure}[!h]
\begin{center}
\includegraphics[width=3.2in]{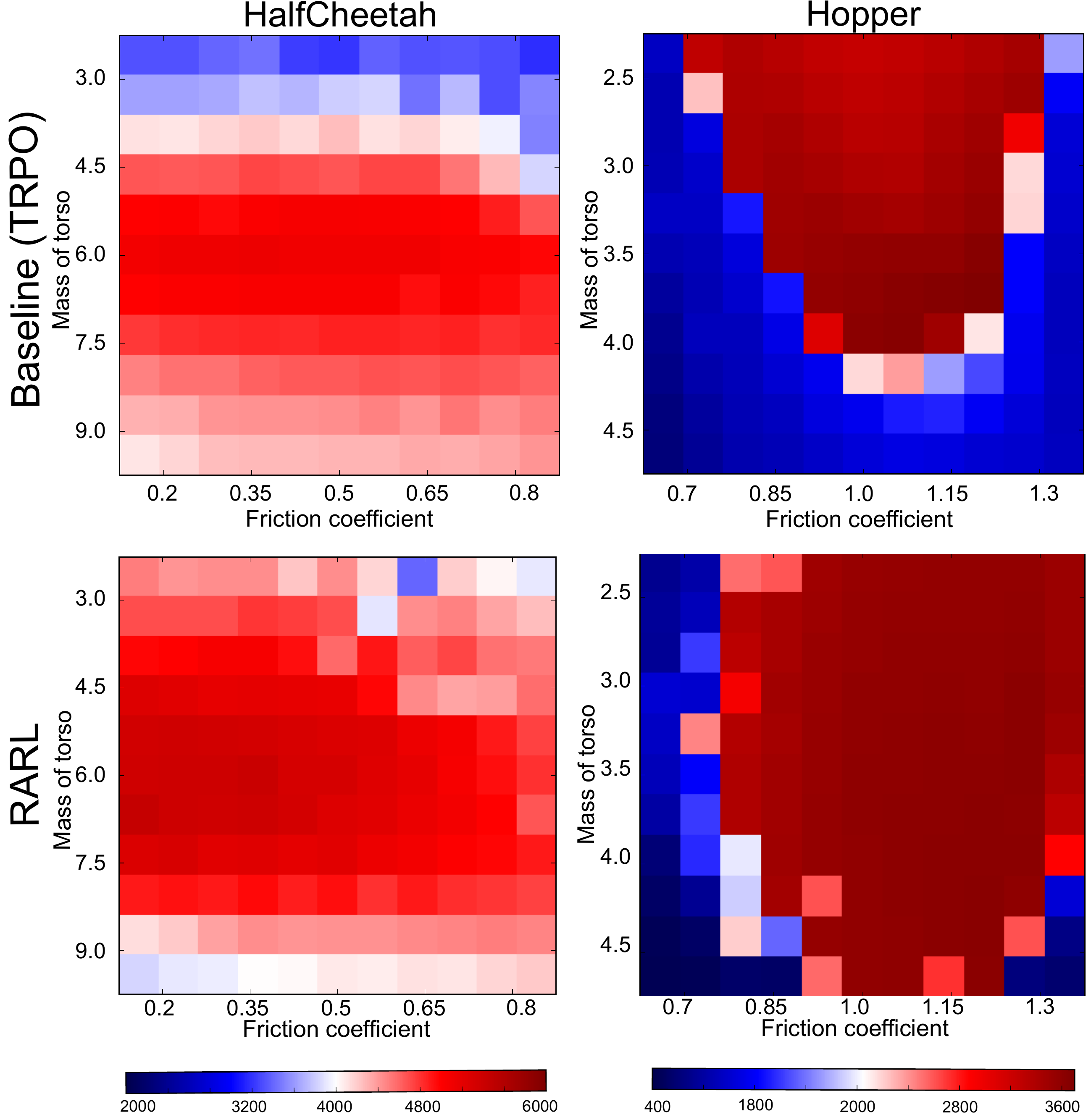}
\end{center}
\caption{The heatmaps show robustness of \textsc{RARL} policies to changing both friction and mass between training and testing. For both the tasks of Hopper and HalfCheetah, we observe a significant increase in robustness.}
\label{fig:rob_joint}
\end{figure}

We visualize the increased robustness of \textsc{RARL} in Figure~\ref{fig:rob_joint}, where we test with jointly varying both mass and friction coefficient. As observed from the figure, for most combinations of mass and friction values RARL leads significantly higher reward values compared to the baseline.

\subsection{Visualizing the Adversarial Policy}
\begin{figure}[h!]
\begin{center}
\includegraphics[width=3.0in]{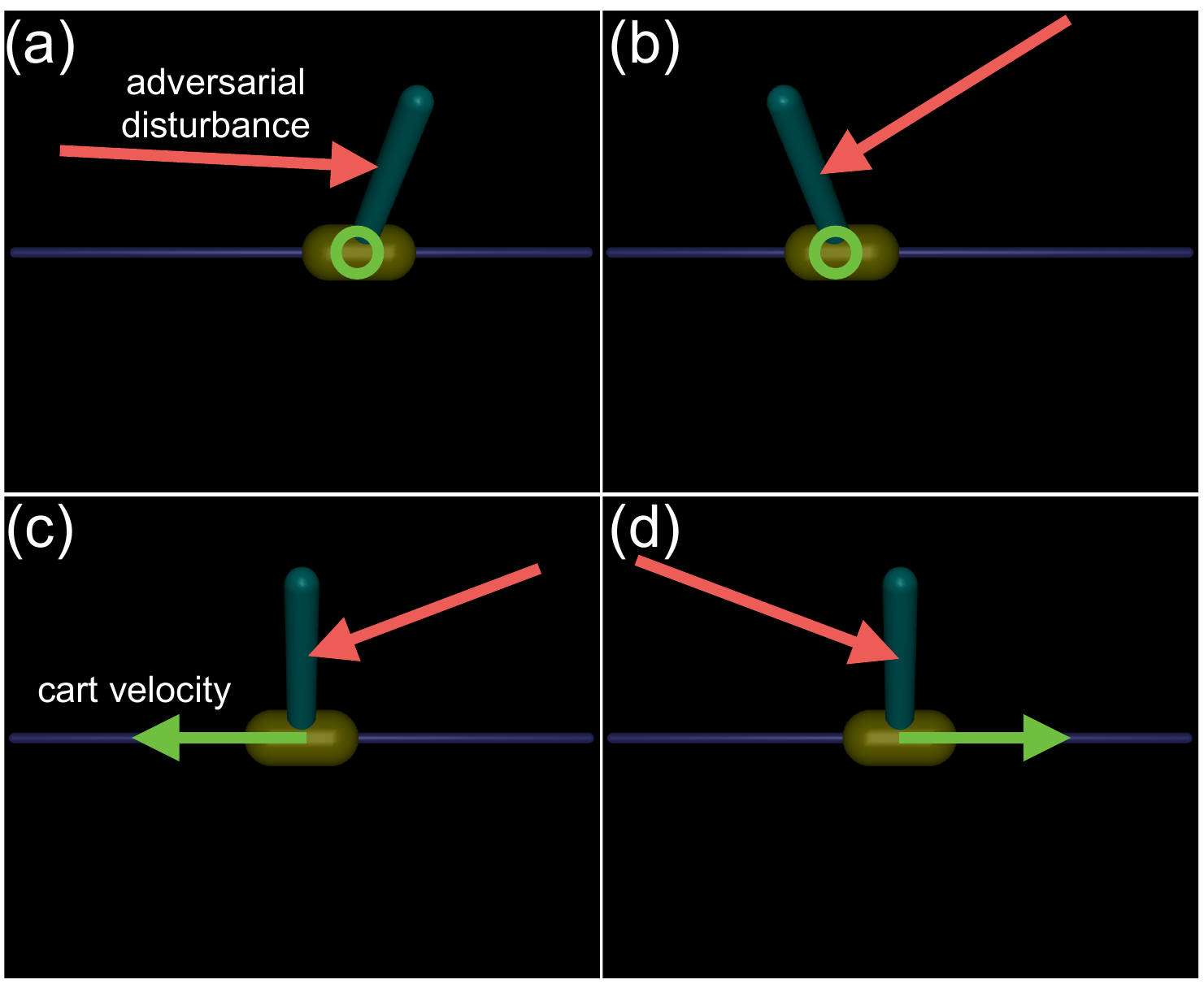}
\end{center}
\caption{Visualization of forces applied by the adversary on InvertedPendulum. In (a) and (b) the cart is stationary, while in (c) and (d) the cart is moving with a vertical pendulum.}
\label{fig:inv_p_adv}
\end{figure}

Finally, we visualize the adversarial policy for the case of InvertedPendulum and Hopper to see whether the learned policies are human interpretable. As shown in Figure~\ref{fig:inv_p_adv}, the direction of the force applied by the adversary agrees with human intuition: specifically, when the cart is stationary and the pole is already tilted (top row), the adversary attempts to accentuate the tilt. Similarly, when the cart is moving swiftly and the pole is vertical (bottom row), the adversary applies a force in the direction of the cart's motion. The pole will fall unless the cart speeds up further (which can also cause the cart to go out of bounds). Note that the naive policy of pushing in the opposite direction would be less effective since the protagonist could slow the cart to stabilize the pole.

\begin{figure}[h!]
\begin{center}
\includegraphics[width=3.0in]{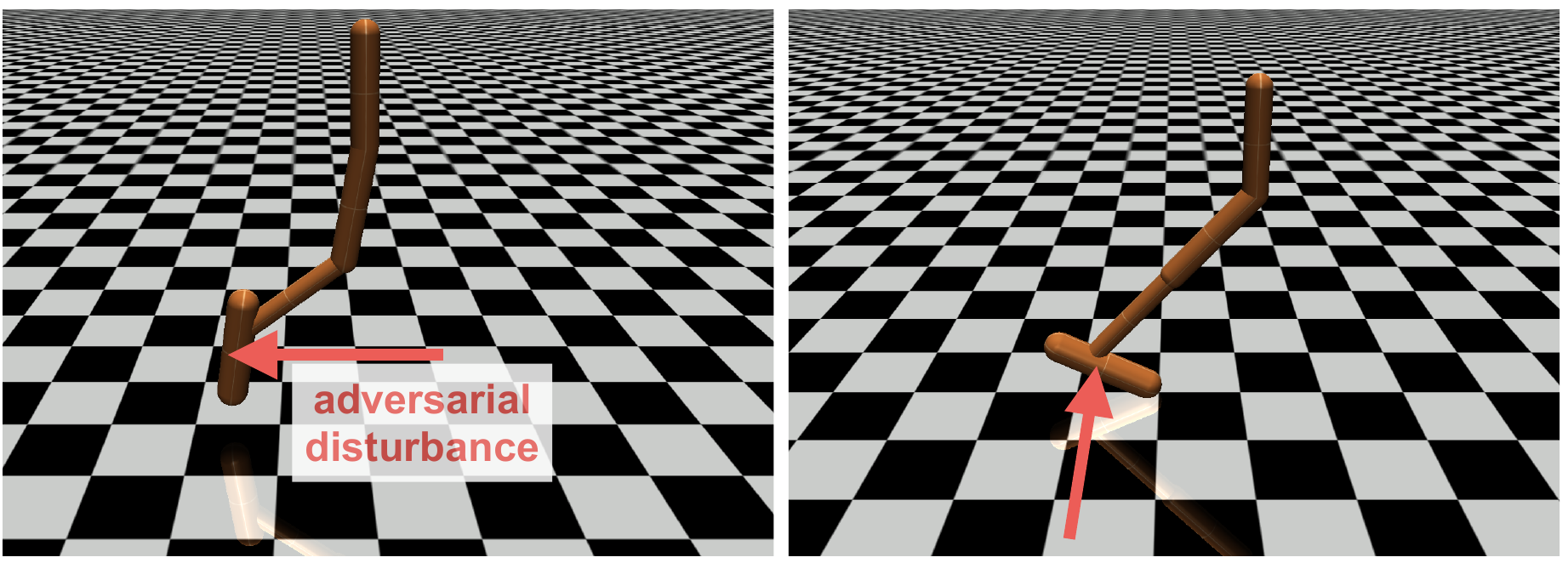}
\end{center}
\caption{Visualization of forces applied by the adversary on Hopper. On the left, the Hopper's foot is in the air while on the right the foot is interacting with the ground.}
\label{fig:hopper_adv}
\end{figure}

Similarly for the Hopper task in Figure~\ref{fig:hopper_adv}, the adversary applies horizontal forces to impede the motion when the Hopper is in the air (left) while applying forces to counteract gravity and reduce friction when the Hopper is interacting with the ground (right).
\section{Related Research}

Recent applications of deep reinforcement learning (deep RL) have shown great success in a variety of tasks ranging from games \cite{mnih2015human, silver2016mastering}, robot control \cite{gu2016continuous, lillicrap2015continuous, mordatch2015interactive}, to meta learning \cite{zoph2016neural}. An overview of recent advances in deep RL is presented in \cite{li2017deep} and \cite{kaelbling1996reinforcement, kober2012reinforcement} provide a comprehensive history of RL research.

Learned policies should be robust to uncertainty and parameter variation to ensure predicable behavior, which is essential for many practical applications of RL including robotics. Furthermore, the process of learning policies should employ safe and effective exploration with improved sample efficiency to reduce risk of costly failure.  These issues have long been recognized and investigated in reinforcement learning \cite{garcia2015comprehensive} and have an even longer history in control theory research \cite{zhou1998essentials}. These issues are exacerbated in deep RL by using neural networks, which while more expressible and flexible, often require significantly more data to train and produce potentially unstable policies.  

In terms of \cite{garcia2015comprehensive} taxonomy, our approach lies in the class of worst-case formulations.  We model the problem as an $H_\infty$ optimal control problem \cite{basar2008h}. In this formulation, nature (which may represent input, transition or model uncertainty) is treated as an adversary in a continuous dynamic zero-sum game. We attempt to find the minimax solution to the reward optimization problem. This formulation was introduced as robust RL (RRL) in \cite{morimoto2005robust}. RRL  proposes  a model-free an actor-disturber-critic method. Solving for the optimal strategy for general nonlinear systems requires is often analytically infeasible for most problems. To address this, we extend RRL's model-free formulation using deep RL via TRPO \cite{schulman2015trust} with neural networks as the function approximator. 

Other worst-case formulations have been introduced. \cite{nilim2005robust} solve finite horizon tabular MDPs using a minimax form of dynamic programming.  Using a similar game theoretic formulation \cite{littman1994markov} introduces the notion of a Markov Game to solve tabular problems, which involves linear program (LP) to solve the game optimization problem. \cite{sharma2007robust} extend the Markov game formulation using a trained neural network for the policy and approximating the game to continue using LP to solve the game. \cite{wiesemann2013robust} present an enhancement to standard MDP that provides probabilistic guarantees to unknown model parameters. Other approaches are risk-based including \cite{tamar2014optimizing, delage2010percentile}, which formulate various mechanisms of percentile risk into the formulation. Our approach focuses on continuous space problems and is a model-free approach that requires explicit parametric formulation of model uncertainty.

Adversarial methods have been used in other learning problems including \cite{goodfellow2014explaining}, which leverages adversarial examples to train a more robust classifiers and \cite{goodfellow2014generative, dumoulin2016adversarially}, which uses an adversarial lost function for a discriminator to train a generative model. In \cite{PintoDG16} two supervised agents were trained with one acting as an adversary for self-supervised learning which showed improved robot grasping. Other adversarial multiplayer approaches have been proposed including \cite{heinrich2016self-play} to perform self-play or fictitious play. Refer to \cite{bucsoniu2010multi} for an review of multiagent RL techniques.

Recent deep RL approaches to the problem focus on explicit parametric model uncertainty. \cite{heess2015memory} use recurrent neural networks to perform direct adaptive control. Indirect adaptive control was applied in \cite{yu2017preparing} for online parameter identification. \cite{Rajeswaran2016epopt} learn a robust policy by sampling the worst case trajectories from a class of parametrized models, to learn a robust policy. 
%

\section{Conclusion}
We have presented a novel adversarial reinforcement learning framework, RARL, that is: (a) robust to training initializations; (b) generalizes better and is robust to environmental changes between training and test conditions; (c) robust to disturbances in the test environment that are hard to model during training. Our core idea is that modeling errors should be viewed as extra forces/disturbances in the system.
Inspired by this insight, we propose modeling uncertainties via an adversary  that applies disturbances to the system. Instead of using a fixed policy, the adversary is reinforced and learns an optimal policy to optimally thwart the protagonist. Our work shows that the adversary effectively samples hard examples (trajectories with worst rewards) leading to a more robust control strategy. 


\bibliography{example_paper}
\bibliographystyle{icml2017}

\end{document}